\documentclass[11pt]{article}
\usepackage{breakcites}
\usepackage{coling2014}
\usepackage{times}
\usepackage{url}
\usepackage{latexsym}
\usepackage{amsmath}
\usepackage{multirow,bigdelim}
\usepackage{amssymb}
\usepackage{graphicx}
\usepackage{xcolor}
\usepackage{enumitem}
\newcommand{\mbf}[1]{\mathbf{#1}}
\newcommand{\comments}[1]{}
\newcommand{\tabincell}[2]{\begin{tabular}{@{}#1@{}}#2\end{tabular}}

\newcommand*{\B}[1]{\ifmmode\bm{#1}\else\textbf{#1}\fi}

\title{Neural Responding Machine for Short-Text Conversation}

\author{Lifeng Shang \ Zhengdong Lu \ Hang Li \\
Noah's Ark Lab\\
Huawei Technologies Co. Ltd.\\
Sha Tin, Hong Kong\\
{\tt \{Shang.Lifeng,Lu.Zhengdong,HangLi.HL\}@huawei.com}\\
}

\date{}

\begin{document}
\maketitle
\begin{abstract}
We propose Neural Responding Machine~(NRM), a neural network-based response generator for Short-Text Conversation. NRM takes the general encoder-decoder framework: it formalizes the generation of response as a decoding process based on the latent representation of the input text, while both encoding and decoding are realized with recurrent neural networks~(RNN). The NRM is trained with a large amount of one-round conversation data collected from a microblogging service. Empirical study shows that NRM can generate grammatically correct and content-wise appropriate responses to over 75\% of the input text, outperforming state-of-the-arts in the same setting, including retrieval-based and SMT-based models.
\end{abstract}

\section{Introduction} 
Natural language conversation is one of the most challenging
artificial intelligence problems, which involves language understanding, reasoning, and the utilization of
common sense knowledge. Previous works in this direction mainly focus on either
rule-based or learning-based methods~\cite{williams2007partially,schatzmann2006survey,misu2012reinforcement,litman2000njfun}. These types of methods
often rely on manual effort in designing rules or automatic training of model with a particular learning algorithm and a small amount of data,
which makes it difficult to develop an extensible open domain conversation system.

Recently due to the explosive growth of microblogging services such as Twitter\footnote{https://twitter.com/.}
and Weibo\footnote{http://www.weibo.com/.},
the amount of conversation data available on the web has tremendously
increased. This makes a data-driven approach to attack the conversation problem~\cite{ji2014information,ritter2011data} possible.
Instead of multiple rounds of conversation, the task at hand, referred to as Short-Text Conversation~(STC), only considers one round
of conversation, in which each round is formed by two short
texts, with the former being an input (referred to as post) from a user and
the latter a response given by the computer. The research on STC may shed light on
understanding the complicated mechanism of natural language conversation.

Previous methods for STC fall into two categories, 1) the retrieval-based method~\cite{ji2014information}, and 2) the statistical machine translation (SMT) based method~\cite{ritter2011data}.
The basic idea of retrieval-based method is to pick a suitable response by ranking the candidate responses with a linear or non-linear
combination of various matching features~(e.g. number of shared words). The main drawbacks of the retrieval-based method are the following

\begin{itemize}[leftmargin=18pt,topsep=1pt]
\setlength \itemsep{-0.1em}
  \item the responses are pre-existed and hard to be customized for the particular text or requirement from the task, e.g., style or attitude.
  \item the use of matching features alone is usually not sufficient for distinguishing positive responses from negative ones, even after time consuming feature engineering. (e.g., a penalty due to mismatched named entities is difficult to be incorporated into the model)
\end{itemize}

The SMT-based method, on the other hand, is generative. Basically it treats the response generation as a translation problem, in which the model is trained on a parallel corpus of post-response pairs. Despite its generative nature, the method is intrinsically unsuitable for response generation, because the responses are not semantically equivalent to the posts as in translation\comments{~(e.g. the post is `here is a gift for you' and the response `thank you')}.
Actually one post can receive responses with completely different content, as manifested through the example in the following figure:
\begin{table}[tbhp]
\centering {
\setlength{\tabcolsep}{5pt}
\begin{tabular}{|c|l|}
\hline
Post & Having my fish sandwich right now
\\ \hline \hline
UserA & For god's sake, it is 11 in the morning
\\
UserB & Enhhhh... sounds yummy
\\
UserC & which restaurant exactly?
\\
\hline
\end{tabular}
} 
\label{tab:motivation_example}
\end{table}

\subsection{Overview} 
\begin{figure}[tbh]
\centering
\includegraphics[width=0.45\linewidth]{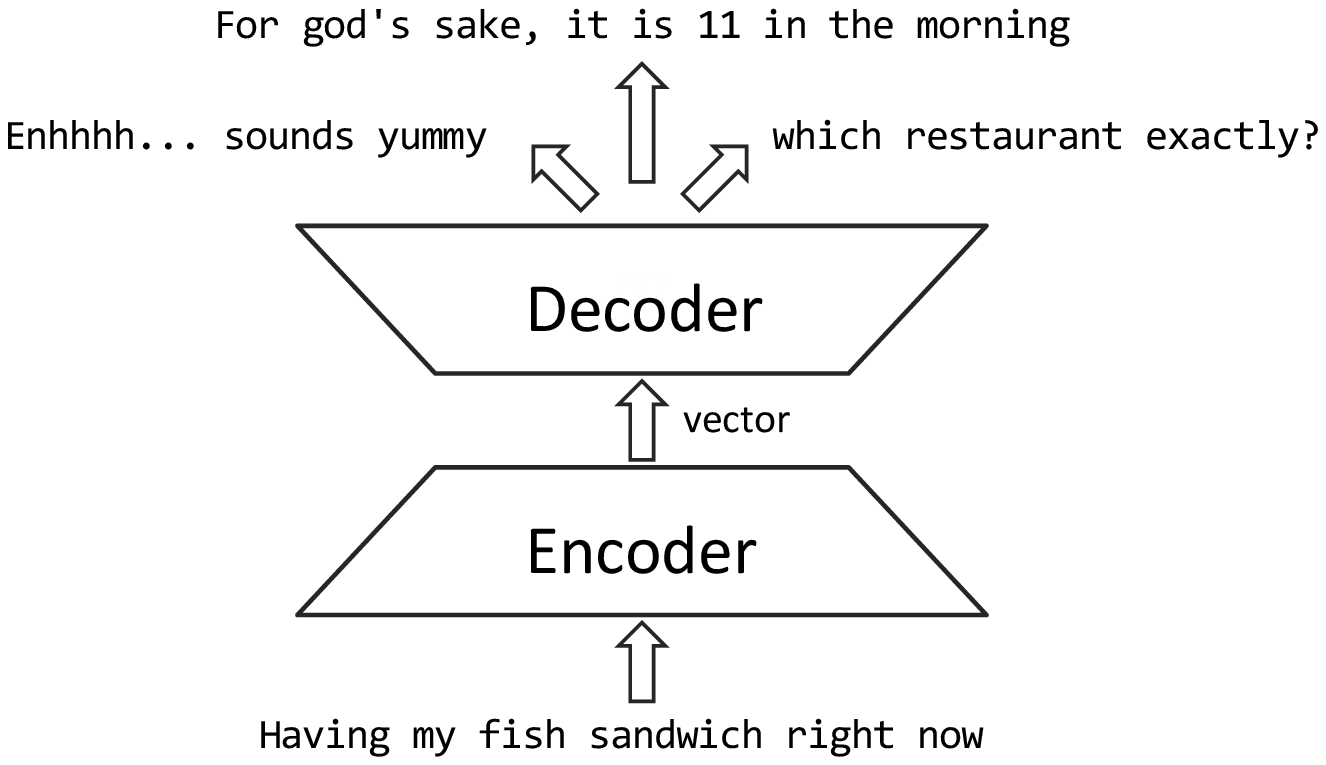}
\caption{The diagram of encoder-decoder framework for automatic response generation.} \label{fig:diagram}
\end{figure}

In this paper, we take a probabilistic model to address the response generation problem, and propose employing a neural encoder-decoder for this task, named \emph{Neural Responding Machine} (NRM). The neural encoder-decoder model, as illustrated in Figure~\ref{fig:diagram}, first summarizes the post as a vector representation, then feeds this representation to decoder to generate responses. We further generalize this scheme to allow the post representation dynamically change during the generation process, following the idea in~\cite{bahdanau2014neural} originally proposed for neural-network-based machine translation with automatic alignment.

NRM essentially estimates the likelihood of a response given a post. Clearly the estimated probability should be complex enough to represent all the suitable responses. Similar framework has been used for machine translation with a remarkable success~\cite{kalchbrenner2013recurrent,auli2013joint,sutskever2014sequence,bahdanau2014neural}. Note that in machine translation, the task is to estimate the probability of a target language sentence conditioned on the source language sentence with the same meaning, which is much easier than the task of STC which we are considering here. In this paper, we demonstrate that NRM, when equipped with a reasonable amount of data, can yield a satisfying estimator
of responses (hence response generator) for STC, despite the difficulty of the task.

Our main contributions are two-folds:
1) we propose to use an encoder-decoder-based neural network to generate a response in STC;
2) we have empirically verified that the proposed method, when trained with a reasonable amount of data, can yield performance better than traditional retrieval-based and translation-based methods.

{\subsection{RoadMap}
In the remainder of this paper, we start with introducing the dataset for STC in Section~\ref{s:dataset}. Then we elaborate on the model of NRM in Section~\ref{s:NRM}, followed by the details on training in Section~\ref{s:training}. After that, we report the experimental results in Section~\ref{s:expts}. In Section~\ref{s:related} we conclude the paper.} 

\section{The Dataset for STC}\label{s:dataset} 
Our models are trained on a corpus of roughly 4.4 million pairs of conversations from Weibo~\footnote{The
dataset and its English translation (by machine translation system) will be released soon.}. 

\subsection{Conversations on Sina Weibo}
Weibo is a popular Twitter-like microblogging service in China, on which a user can post short messages (referred to as \emph{post} in
the reminder of this paper) visible to the public or a group of users following her/him. Other users make comment on a published post, which will be referred to as
\emph{response}.\comments{An example of post and its associated responses is given in Figure~\ref{fig:annotation_examples}.} Just like Twitter, Weibo also has the length limit of 140 Chinese characters on both posts and responses, making the post-response pair an ideal surrogate for short-text conversation. 

\begin{table}[tbhp]
    \centering
        \begin{tabular}{|c|l|r|}
        \hline
        \multirow{3}{*}{\bf Training} & \#posts & 219,905  \\ \cline{2-3}
        \multicolumn{1}{|l|}{}                & \#responses & 4,308,211 \\ \cline{2-3}
        \multicolumn{1}{|l|}{}                & \#pairs & 4,435,959 \\ \hline
        \bf Test Data                             & \#test posts & 110      \\ \hline \hline

        \multirow{3}{*}{\tabincell{c}{\bf Labeled Dataset \\ \small (retrieval-based)}} & \#posts & 225  \\ \cline{2-3}
        \multicolumn{1}{|l|}{}                & \#responses & 6,017 \\ \cline{2-3}
        \multicolumn{1}{|l|}{}                & \#labeled pairs & 6,017 \\ \hline \hline

        \multirow{3}{*}{\tabincell{c}{\bf Fine Tuning \\ \small (SMT-based)}} & \#posts & 2,925  \\ \cline{2-3}
        \multicolumn{1}{|l|}{}                & \#responses & 3,000 \\ \cline{2-3}
        \multicolumn{1}{|l|}{}                & \#pairs & 3,000 \\ \hline

        \end{tabular}
    \caption{Some statistics of the dataset. {\bf Labeled Dataset} and {\bf Fine Tuning} are used by retrieval-based method for learning to rank and SMT-based method for fine tuning, respectively. }\label{tab:statistic_dataset}
    \end{table}

\subsection{Dataset Description} 
To construct this million scale dataset, we first crawl hundreds of millions of post-response pairs, and then clean the raw data in a similar way as suggested in~\cite{wang2013dataset}, including 1) removing trivial responses like ``wow'', 2) filtering out potential advertisements, and 3) removing the responses after first 30 ones for topic consistency.  Table~\ref{tab:statistic_dataset} shows some
statistics of the dataset used in this work. It can be seen that each post have 20 different responses on average. In addition to the semantic gap between post and its responses,
this is another key difference to a general parallel data set used for traditional translation.

\section{Neural Responding Machines for STC}\label{s:NRM} 
\begin{figure}[!ht]
\centering
\includegraphics[width=0.35\linewidth]{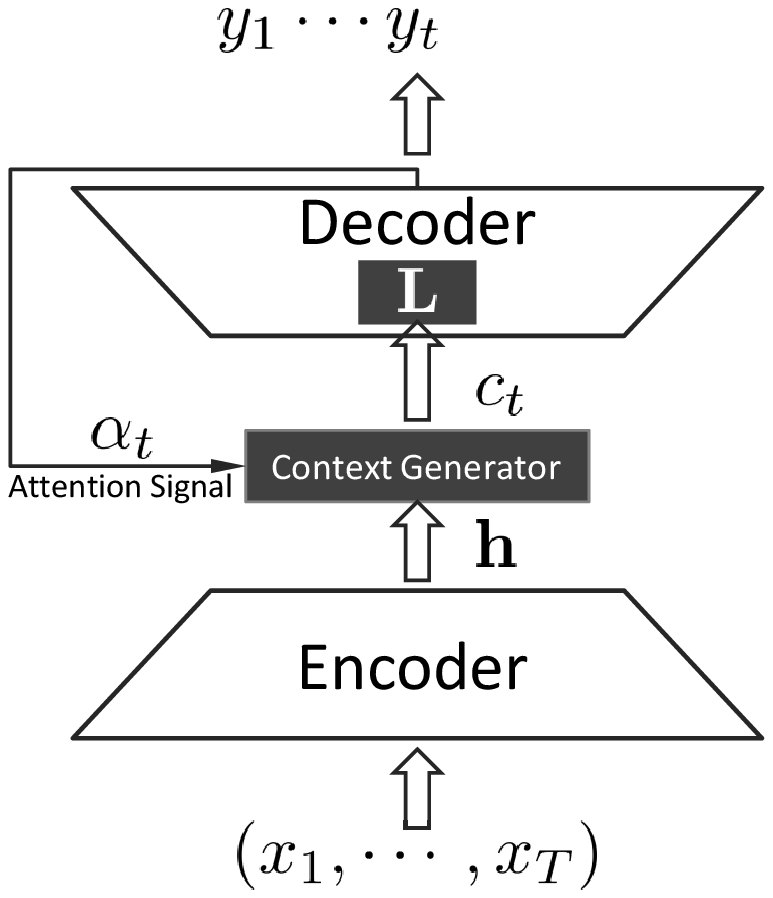}
\caption{The general framework and dataflow of the encoder-decoder-based NRM.} \label{fig:fw-enc-dec}
\end{figure}

The basic idea of NRM is to build a hidden representation of a post, and then generate the response based on it, as shown in Figure~\ref{fig:fw-enc-dec}. In the particular illustration, the encoder converts the input sequence $\mbf{x}=(x_{1}, \cdots, x_{T})$
into a set of high-dimensional hidden representations $\mbf{h}=(h_{1}, \cdots, h_{T})$,  which, along with the attention signal at time $t$ (denoted as $\alpha_t$), are fed to the context-generator to build the context input to decoder at time $t$ (denoted as $c_{t}$). Then $c_{t}$ is linearly transformed by a matrix $\mathbf{L}$ (as part of the decoder)
into a stimulus of generating RNN to produce the $t$-th word of response~(denoted as $y_t$).

In neural translation system, $\mathbf{L}$ converts the representation in source language to that of target language. In NRM, $\mathbf{L}$ plays a more difficult role: it needs to transform the representation of post (or some part of it) to the rich representation of many plausible responses.  It is a bit surprising that this can be achieved to a reasonable level with a linear transformation in the ``space of representation", as validated in Section~\ref{s:case}, where we show that one post can actually invoke many different responses from NRM.

The role of attention signal is to determine which part of the
hidden representation $\mbf{h}$ should be emphasized during the generation process. It should be noted that $\alpha_t$ could be fixed over time or changes dynamically during the generation of response sequence $\mbf{y}$. In the dynamic settings, $\alpha_t$ can be function of historically generated subsequence $(y_1, \cdots, y_{t-1})$, input sequence $\mbf{x}$ or their latent representations, more details will be shown later in Section~\ref{s:encoder}.

We use Recurrent Neural Network~(RNN) for both encoder and decoder, for its natural ability to summarize and generate word sequence of arbitrary lengths~\cite{mikolov2010recurrent,sutskever2014sequence,cho2014learning}.

\begin{figure}[tbh]
\centering
\includegraphics[width=0.35\linewidth]{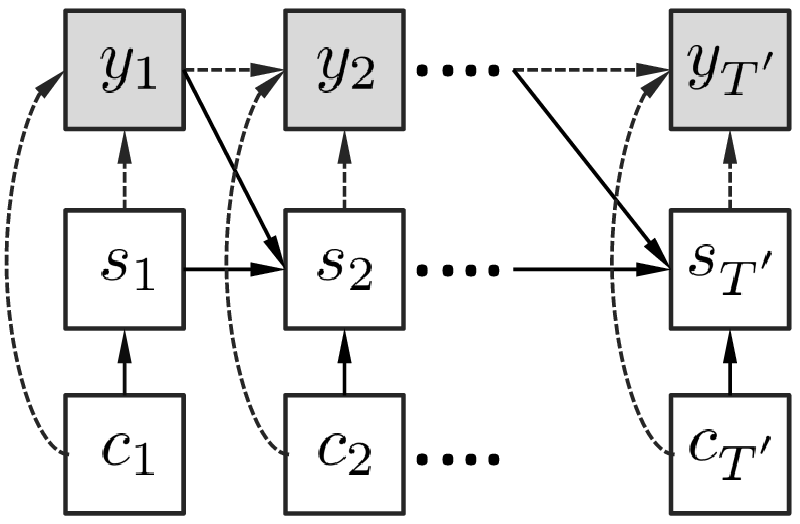}
\caption{The graphical model of RNN decoder. The dashed lines denote the
variables related to the function $g(\cdot)$, and the solid lines denote the variables
related to the function $f(\cdot)$.} \label{fig:decoder}
\end{figure}

\subsection{The Computation in Decoder} 
{\color{black} Figure~\ref{fig:decoder} gives the graphical model of the decoder, which is essentially a standard RNN language model except conditioned on the context input $\mbf{c}$.
The generation probability of the $t$-th word is calculated by
\begin{align}
  p(y_t|y_{t-1}, \cdots, y_{1}, \mbf{x}) = g(y_{t-1}, s_t, c_t),
\end{align}
where $y_t$ is a one-hot word representation, $g(\cdot)$ is a softmax activation function, and $s_t$ is the hidden state of decoder at time $t$ calculated by
\begin{align}
  s_t = f(y_{t-1}, s_{t-1}, c_t),
\end{align}
and $f(\cdot)$ is a non-linear activation function and the transformation $\mathbf{L}$ is often assigned as parameters of $f(\cdot)$.} Here $f(\cdot)$ can be a logistic function, the sophisticated long short-term memory (LSTM) unit~\cite{hochreiter1997long}, or the recently proposed gated recurrent unit~(GRU)~\cite{chung2014empirical,cho2014learning}. Compared to ``ungated" logistic function, LSTM and GRU are specially designed for its long term memory: it can store information over extended time steps without too much decay.  We use GRU in this work,  since it performs comparably to LSTM on squence modeling~\cite{chung2014empirical}, but has less parameters and easier to train.

\subsection{The Computation in Encoder}\label{s:encoder} 
\begin{figure}[!htbp]
\centering
\includegraphics[width=0.35\linewidth]{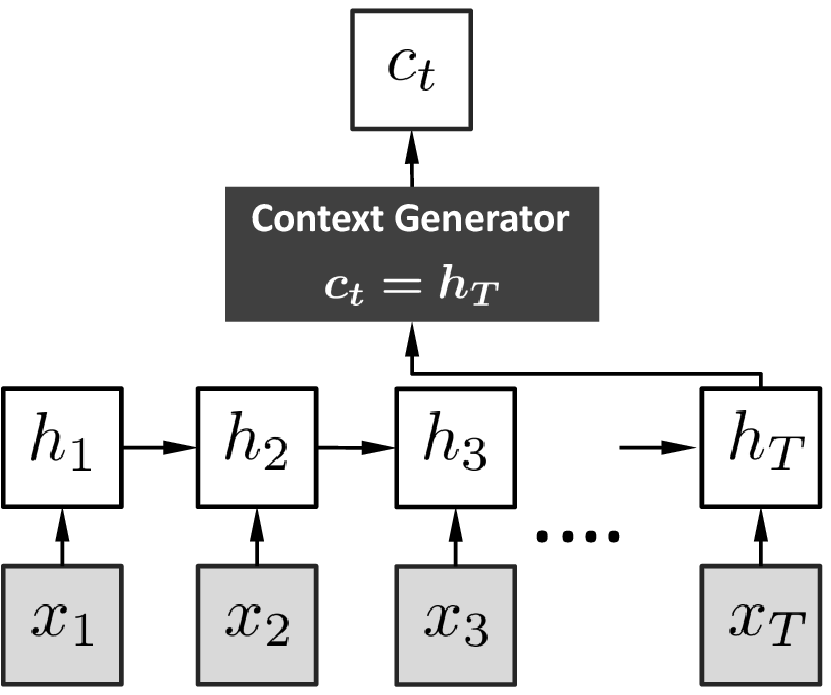}
\caption{The graphical model of the encoder in NRM-glo, where the last hidden state is used as the context vector $c_t = h_T$.} \label{fig:encoder-glo}
\end{figure}

We consider three types of encoding schemes, namely 1) the global scheme, 2) the local scheme, and the hybrid scheme which combines 1) and 2).
\paragraph{Global Scheme:} Figure~\ref{fig:encoder-glo} shows the graphical model of the RNN-encoder and related context generator for a global encoding scheme. The hidden state at time $t$ is calculated by $h_t = f(x_t, h_{t-1})$~(i.e. still GRU unit),
\comments{
\vspace{-0.2cm}
\begin{align}
  h_t = f(x_t, h_{t-1}),
\end{align}
}and with a trivial context generation operation, we essentially use the final hidden state $h_T$ as the global representation of the sentence. The same strategy has been taken in \cite{cho2014learning} and \cite{sutskever2014sequence} for building the intermediate representation for machine translation. This scheme however has its drawbacks: a vectorial summarization of the entire post is often hard to obtain and may lose important details for response generation, especially when the dimension of the hidden state is not big enough\footnote{\newcite{sutskever2014sequence} has to use $4,000$ dimension for satisfying performance on machine translation, while \cite{cho2014learning} with a smaller dimension perform poorly on translating an entire sentence.}. In the reminder of this paper, a NRM with this global encoding scheme is referred to as NRM-glo. 

\begin{figure}[!htbp]
\centering
\includegraphics[width=0.35\linewidth]{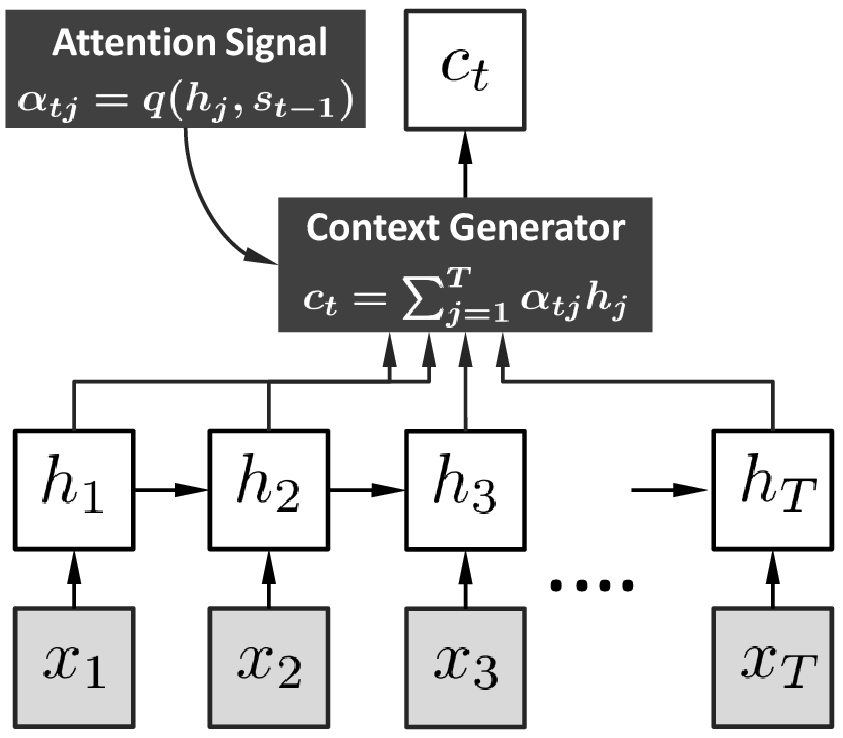}
\caption{The graphical model of the encoder in NRM-loc, where the weighted sum of hidden sates is used as the context vector $c_t = \sum_{j=1}^{T}\alpha_{tj}h_j$.} \label{fig:encoder-loc}
\end{figure}

\paragraph{Local Scheme:} Recently, \newcite{bahdanau2014neural} and \newcite{graves2013generating} introduced an attention mechanism that allows the decoder to dynamically select and linearly combine different parts of the input sequence $c_t = \sum_{j=1}^{T}\alpha_{tj}h_j$,
\comments{
\begin{align}
  c_t = \sum_{j=1}^{T}\alpha_{tj}h_j,
  \label{eqn:attention_sum}
\end{align} }
where weighting factors $\alpha_{tj}$ determine which part should be selected to generate the new word $y_t$, which in turn is a function of hidden states $\alpha_{tj} = q(h_j, s_{t-1})$, as pictorially shown in Figure~\ref{fig:encoder-loc}. Basically, the attention mechanism $\alpha_{tj}$ models the alignment between the inputs around position $j$ and the output at position $t$, so it can be viewed as a local matching model. This local scheme is devised in~\cite{bahdanau2014neural} for automatic alignment between the source sentence and the partial target sentence in machine translation.
{\color{black}This scheme enjoys the advantage of adaptively focusing on some important words of the input text according to the generated words of response.}
A NRM with this local encoding scheme is referred to as NRM-loc. 
\comments{By this strategy the linear transformation $\mathbf{T}$ can be also freed from context selection to some extent. The structural knowledge of the input sentence can also be incorporated through $\alpha_{tj}$, e.g. placing relatively higher weights to the ancestor nodes of the input syntax tree.}

\subsection{Extensions: Local and Global Model}
\begin{figure}[!tbh]
\centering
\includegraphics[width=0.45\linewidth]{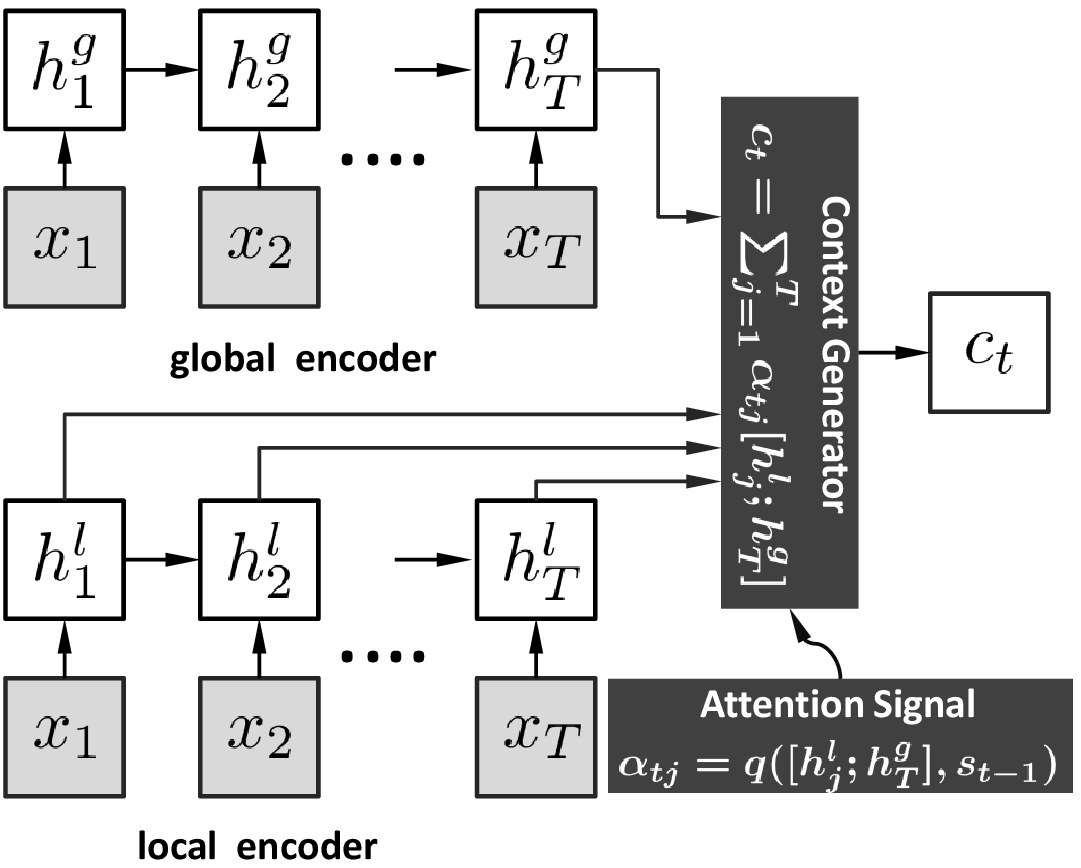}
\caption{The graphical model for the encoder in NRM-hyb, while context generator function is $c_t=\sum_{j=1}^{T}\alpha_{tj}[h_j^l; h_T^g]$, here $[h_j^l; h_T^g]$ denotes the concatenation of
vectors $h_j^l$ and $h_T^g$} \label{fig:encoder-hyb}
\end{figure}

In the task of STC, NRM-glo has the summarization of the entire post, while NRM-loc can adaptively select the important words in post for various suitable responses. Since post-response pairs in STC are not strictly parallel and a word in different context can have different meanings, we conjecture that the global representation in NRM-glo may provide useful context for extracting the local context, therefore complementary to the scheme in NRM-loc. It is therefore a natural extension to combine the two models by concatenating their encoded hidden states to form an extended hidden representation for each time stamp, as illustrated in Figure~\ref{fig:encoder-hyb}. We can see the summarization $h_T^g$ is incorporated into $c_t$ and $\alpha_{tj}$ to provide a global context for local matching. With this hybrid method, we hope both the local and global information can be introduced into the generation of response. The model with this context generation mechanism is denoted as NRM-hyb.

\comments{\color{black} It should be noted that different context generation mechanisms will result the difference in representability of hidden states. Specifically, the last hidden state of NRM-glo should be different from that of the NRM-loc, since it has the stress to encode all information of the input sentence. In the training stage of NRM-hyb, we use the trick of initializing NRM-hyb with the learned parameters of NRM-loc and NRM-glo, then fine tuning the parameters in encoder and normally adjusting the parameters in decoder.
}

It should be noticed that the context generator in NRM-hyb will evoke different encoding mechanisms in the global encoder and the local encoder, although they will be combined later in forming a unified representation. More specifically, the last hidden state of NRM-glo plays a role different from that of the last state of NRM-loc, since it has the responsibility to encode the entire input sentence. This role of NRM-glo, however, tends to be not adequately emphasized in training the hybrid encoder when the parameters of the two encoding RNNs are learned jointly from scratch. For this we use the following trick: we first initialize NRM-hyb with the parameters of NRM-loc and NRM-glo trained separately, then fine tune the parameters in encoder along with training the parameters of decoder.

\comments{
\begin{algorithm}                      
\caption{Training algorithm of NRM-hyb.}          
\label{alg1}                           
\begin{algorithmic}                    

\REQUIRE The post-response pairs $(\mbf{x}, \mbf{y})\in \mathcal{S}$. The learned parameters
of NRM-${glo}$ and NRM-loc.

\ENSURE  The learned parameters of encoder $\Theta^{hyb}$ and decoder $\Psi^{hyb}$

\medskip
(Step 1) Training models NRM-loc and NRM-glo

\STATE Training the NRM-${loc}$ and get the parameters $\Theta^{loc}$ and $\Psi^{loc}$.
\STATE $s(1) = \arg \max_{n} H(A_{a}^{n})$

\STATE $A_{a}^{S} = \{A_{a}^{s(1)}\}$

\STATE $A_{a}^{C} \leftarrow A_{a} - A_{a}^{S}$
\medskip

(Step 2) Training the NRM-${glo}$.

\FOR{$k$ = 1 to $(K - 1)$}

\STATE $s(k + 1) = \arg \max_{n}\{ \min_{l \leq k}H(A_{a}^{n} |
A_{a}^{s(l)})\}$, $A_{a}^{n} \in A_{a}^{C}$

\STATE $A_{a}^{S} \leftarrow A_{a}^{S} \cup \{A_{a}^{s(k+1)}\}$
\medskip

\STATE $A_{a}^{C} \leftarrow A_{a} - A_{a}^{S}$

\ENDFOR
\medskip
\end{algorithmic}
\end{algorithm}
}

To learn the parameters of the model, we maximize the likelihood of observing the original response conditioned on the post in the training set.  For a new post, NRMs generate their responses by using a left-to-right beam search with beam size = 10.

\section{Experiments}\label{s:training} 
We evaluate three different settings of NRM described in Section~\ref{s:NRM}, namely NRM-glo, NRM-loc, and NRM-hyb, and compare them to retrieval-based and SMT-based methods. 

\comments{

\begin{table}[h]
\begin{center}
\begin{tabular}{|l|l|}
\hline \bf Name & \bf Description \\ \hline \hline
\multirow{3}{*}{NRM-glo} & The neural model by using the \\
& last hidden state $h_{T}$ as the con-\\
& text input. \\ \hline

\multirow{3}{*}{NRM-loc} & The neural model by using the\\
& weighted sum of hidden states   \\
& as the context input\\ \hline

\multirow{3}{*}{NRM-hyb} & The neural model by using the\\
& hybrid of the encoders of NR- \\
& M$^{glo}$ and NRM$^{loc}$ \\ \hline

\multirow{3}{*}{IR-$l2r$} & IR-based method with fourteen\\
& different features and learned  \\
& by learning to rank  \\ \hline

\multirow{3}{*}{SMT} & Statistical machine translation \\
& model by using the Moses with \\
& default settings. \\
\hline
\end{tabular}
\end{center}
\caption{Summary of compared methods. }
\label{tab:model_summary}
\end{table}

}
\subsection{Implementation Details} 
We use Stanford Chinese word segmenter~\footnote{http://nlp.stanford.edu/software/segmenter.shtml} to split the posts and responses into sequences of words. Although both posts and responses are written in the same language, the distributions on words for the two are different: the number of unique words in post text is 125,237, and that of response text is 679,958. We therefore construct two separate vocabularies for posts and responses by using 40,000 most frequent words on each side, covering 97.8\% usage of words for post and 96.2\% for response respectively. All the remaining words are replaced by a special token ``\texttt{UNK}''. \comments{\color{blue} The average appearance time of the ``UNK'' words is 15.7 in post and reduces to 2.7 in response.}
The dimensions of the hidden states of encoder and decoder are both 1,000, and the dimensions of the word-embedding for post and response are both 620. Model parameters are initialized by randomly sampling from a uniform distribution between -0.1 and 0.1. All our models were trained on a NVIDIA Tesla K40 GPU using stochastic gradient descent algorithm with mini-batch. The training stage of each model took about two weeks. \comments{\color{black}In the training stage of NRM-hyb, we use the trick of initializing NRM-hyb with the learned parameters of NRM-loc and NRM-glo, then fine tuning the parameters in encoder and normally adjusting the parameters in decoder.}

\subsection{Competitor Models} 
\paragraph{Retrieval-based:} with retrieval-based models, for any given post $p^{*}$, the response $r^{*}$ is  retrieved from a big post-response pairs $(p, r)$ repository. Such models rely on three key components: a big
repository, sets of feature functions $\Phi_i(p^{*}, (p, r))$, and a machine learning model to combine these features. In this work, the whole 4.4 million
Weibo pairs are used as the repository, 14 features, ranging from simple cosine similarity to some deep matching models~\cite{ji2014information} are used to determine the suitability of a post to a given post $p^{*}$ through the following linear model
\begin{align}
 score(p^{*}, (p, r)) = \sum_{i}\omega_i\Phi_i(p^{*}, (p, r)).
\end{align}
Following the ranking strategy in~\cite{ji2014information}, we pick 225 posts and about 30 retrieved responses for each of them given by a baseline retriever\footnote{we use the default similarity function of Lucene~\footnote{http://lucene.apache.org/}} from the 4.4M repository, and manually label them to obtain \emph{labeled} 6,017 post-response pairs.
We use ranking SVM model~\cite{joachims2006training} for the parameters $\omega_i$ based on the labeled dataset. In comparison to NRM, only the top one response is considered in the evaluation process.

\paragraph{SMT-based:}
In SMT-based models, the post-response pairs are directly used as parallel data for training a translation model. We use the most widely used open-source phrase-based translation model-Moses~\cite{koehn2007moses}. Another parallel data consisting of 3000 post-response pairs is used to tune the system. In~\cite{ritter2011data}, the authors used a modified SMT model to obtain the ``Response'' of Twitter ``Stimulus''. The main modification is in replacing the standard GIZA++ word alignment model~\cite{och2003systematic} with a new phrase-pair selection method, in which all the possible phrase-pairs in the training data are considered and their associated probabilities are estimated by the Fisher's Exact Test, which yields performance slightly better than default setting\footnote{Reported results showed that the new model outperformed the baseline SMT model 57.7\% of the time.}.
Compared to retrieval-based methods, the generated responses by SMT-based methods often have fluency or even grammatical problems. In this work, we choose the Moses with default settings as our SMT model. 

\begin{figure*}[!htb]
\centering
\includegraphics[width=1\linewidth]{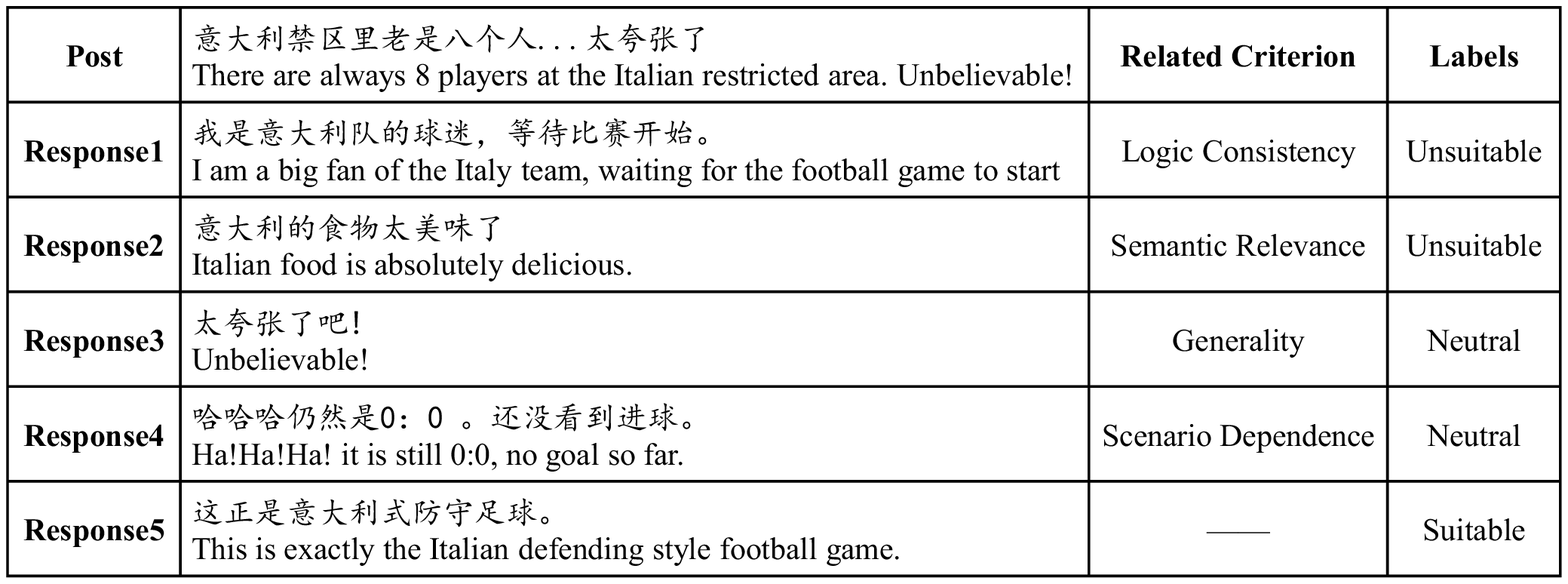}
\caption{An example post and its five candidate responses with human annotation. {The content of the post implies that the football match is already started, while the author of Response1 is still waiting for the match to start. Response2 talks about the food of Italy. Response3 is a widely used response, but it is suitable to this post. Response4 states that the current score is still 0:0, it is a suitable response only in this specific scenario.} \comments{\color{blue}[may cut it if short on space]}} \label{fig:annotation_examples}
\end{figure*}

\section{Results and Analysis}\label{s:expts} 
Automatic evaluation of response generation is still an open problem. The widely accepted evaluation methods in translation~(e.g. BLUE score~\cite{papineni2002bleu}) do not apply, since the range of the suitable responses is so large that it is practically impossible to give reference with adequate coverage. It is also not reasonable to evaluate with Perplexity, a generally used measurement in statistical language modeling, because the naturalness of response and the relatedness to post can not be well evaluated.\comments{\color{blue} [need to rephrase]} We therefore resort to human judgement, similar to that taken in~\cite{ritter2011data} but with important difference. 

\begin{table*}[tbhp]
\centering {
\begin{tabular}{|l|c|c|c|c||c|}
\hline
\bf Models & \bf Mean Score & \bf Suitable (+2) & \bf Neutral (+1) & \bf Unsuitable (0) & \bf  Agreement \\ \hline
 NRM-glo & 0.969 & 34.0\% & 28.9\% & 37.1\% & 0.397\\ \hline
 NRM-loc & 1.065 & 36.0\% & 34.5\% & 29.5\% & 0.247\\ \hline
 NRM-hyb & 1.156 & 39.3\% & 37.1\% & 23.6\% & 0.309\\ \hline \hline
 Rtr.-based & 0.984  & 29.8\% & 38.7\% & 31.5\% & 0.346\\ \hline \hline
 SMT-based & 0.313 &  5.6 \% & 20.0\% & 74.4\% & 0.448\\ \hline
\end{tabular}
}
\caption{The results of evaluated methods. Mean score is the average value of annotated scores over all annotations. \comments{The columns {\bf Suitable (+2)}, {\bf Neutral (+1)}, and {\bf Unsuitable (0)} list the fraction of annotations with annotated scores +2, +1, and 0 respectively. The column {\bf Agreement} lists the Fleiss' kappa value for each method.} (Rtr.-based means the retrieval-based method)}
\label{tab:result_score}
\end{table*}

\comments{
\begin{table}[tbhp]
\centering {
\begin{tabular}{|l|l|c|c|}
\hline
 \bf Model A & \bf Model B & \bf  A $>$ B & \bf A $=$ B\\ \hline
 NRM-loc & NRM-glo & 47.3 \% & 14.6 \% \\ \hline
 NRM-hyb & NRM-glo & 50.9 \% & 14.6 \% \\ \hline
 NRM-hyb & NRM-loc & 45.5 \% & 17.3 \% \\ \hline \hline

 IR-based & NRM-glo & 45.4 \% & 7.3 \% \\ \hline
 IR-based & NRM-loc & 35.5 \% & 10.9 \% \\ \hline
 IR-based & NRM-hyb & 40.0 \% & 7.3 \% \\ \hline \hline

 SMT & NRM-hyb & 12.7 \% & 6.4 \%  \\ \hline
 SMT & IR-based & 11.8 \% & 8.2 \%   \\ \hline
\end{tabular}
} \caption{The comparison between different methods.}
\label{tab:result_compare}
\end{table}
}

\begin{figure*}[!tbh]
\centering
\includegraphics[width=1\linewidth]{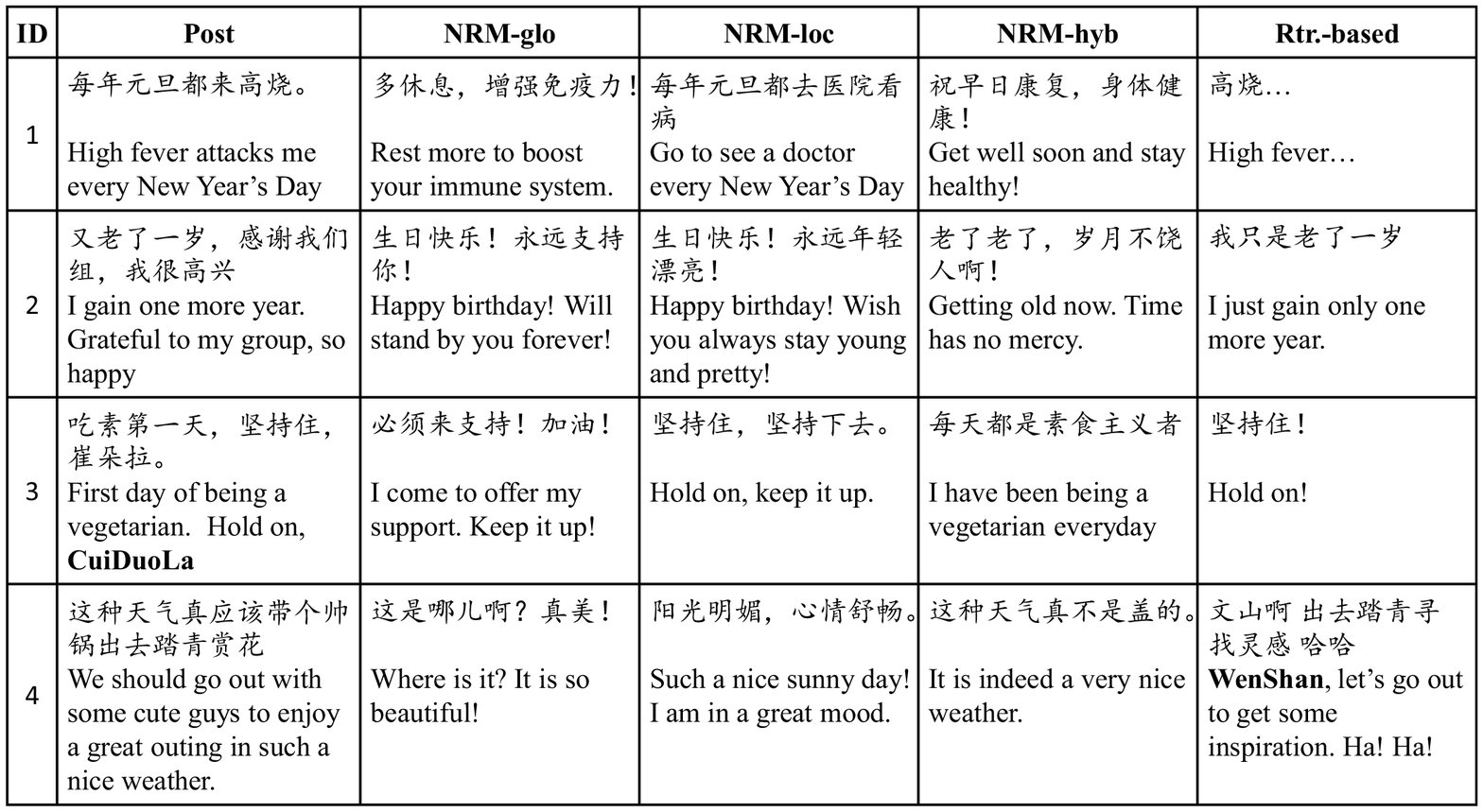}
\caption{Some responses generated by different models (originally in Chinese with their literal English translation), where the words in boldfaces are entity names. } \label{fig:response_examples}
\end{figure*}

\subsection{Evaluation Methods} 
We adopt human annotation to compare the performance of different models. Five labelers with at least three-year experience of Sina Weibo are invited to do human evaluation. Responses obtained from the five evaluated models are pooled and randomly permuted for each labeler. The labelers are instructed to imagine that they were the authors of the original posts and judge whether a response (generated or retrieved) is appropriate and natural to a input post. Three levels are assigned to a response with scores from 0 to 2:
\begin{itemize}[leftmargin=18pt,topsep=1pt]
\setlength \itemsep{-0.1em}
  \item {\bf Suitable (+2):} the response is evidently an appropriate and natural response to the post;
  \item {\bf Neutral (+1):} the response can be a suitable response in a \emph{specific} scenario;
  \item {\bf Unsuitable (0):} it is hard or impossible to find a scenario where response is suitable.
\end{itemize}

To make the annotation task operable, the suitability of generated responses is judged from the following five criteria:
\begin{itemize}[leftmargin=18pt,topsep=1pt]
  \setlength \itemsep{-0.1em}
  \item[(a)] {\bf Grammar and Fluency}: Responses should be natural language and free of any fluency or grammatical errors;
  \item[(b)] {\bf Logic Consistency}: Responses should be logically consistent with the test post;
  \item[(c)] {\bf Semantic Relevance}: Responses should be semantically relevant to the test post;
  \item[(d)] {\bf Scenario Dependence}: Responses can depend on a specific scenario but should not contradict the first three criteria;
  \item[(e)] {\bf Generality}: Responses can be general but should not contradict the first three criteria;
\end{itemize}
If any of the first three criteria (a), (b), and (c) is contradicted, the generated response should be labeled as ``Unsuitable''. The responses that are general or suitable to post in a specific scenario should be labeled as ``Neutral''. Figure~\ref{fig:annotation_examples} shows an example of the labeling results of a post and its responses. The first two responses are labeled as ``Unsuitable'' because of the logic consistency and semantic relevance errors. Response4 depends on the scenario (i.e., the current score is 0:0), and is therefore annotated as ``Neutral". 
\comments{30 posts have the ``UNK'' word and the total number of these out-of-vocabulary words is 54.}

\begin{table}[tbhp]
\centering {
\begin{tabular}{|l|l|c|c|}
\hline
 \bf Model A & \bf Model B & \bf  \tabincell{c}{Average \\ rankings} & \bf $p$ value\\ \hline
 {\bf NRM-loc} & NRM-glo & (1.463, 1.537) & 2.01\% \\ \hline
 {\bf NRM-hyb} & NRM-glo & (1.434, 1.566) & 0.01\% \\ \hline
 {\bf NRM-hyb} & NRM-loc & (1.465, 1.535) & 3.09\% \\ \hline \hline
 Rtr.-based &  NRM-glo & (1.512, 1.488) &  48.1\% \\ \hline
 Rtr.-based &  NRM-loc & (1.533, 1.467) &  6.20\% \\ \hline
 Rtr.-based & {\bf NRM-hyb} & (1.552, 1.448) & 0.32\% \\ \hline \hline
  SMT & {\bf NRM-hyb} & (1.785, 1.215) & 0.00 \%  \\ \hline
 SMT & {\bf Rtr.-based} &  (1.738, 1.262) & 0.00 \%   \\ \hline
\end{tabular}
} \caption{$p$-values and average rankings of Friedman test for
pairwise model comparison. (Rtr.-based means the retrieval-based method)}
\label{tab:result_compare}
\end{table}
\subsection{Results}
Our test set consists of 110 posts that do not appear in the training set, with length between 6 to 22 Chinese words and 12.5 words on average. The experimental results based on human annotation are summarized in Table~\ref{tab:result_score}, consisting of the ratio of three categories and the agreement among the five labelers for each model. The agreement is evaluated by Fleiss' kappa~\cite{fleiss1971measuring}, as a statistical measure of inter-rater consistency.
Except the SMT-based model, the value of agreement is in a range from 0.2 to 0.4 for all the other models, which should be interpreted as ``Fair agreement''. The SMT-based model has a relatively higher kappa value 0.448, which is larger than 0.4 and considered as ``Moderate agreement'', since the responses generated by the SMT often have the fluency and grammatical errors, making it easy to reach an agreement on such unsuitable cases.

From Table~\ref{tab:result_score}, we can see the SMT method  performs significantly worse than the retrieval-based and NRM models and 74.4\% of the generated responses were labeled as unsuitable mainly due to fluency and relevance errors. This observation confirms with our intuition that the STC dataset, with one post potentially corresponding to many responses, can not be simply taken as parallel corpus in a SMT model.  Surprisingly, more than 60\% of responses generated by all the three NRM are labeled as ``Suitable'' or ``Neutral'', which means that most generated responses are fluent and semantically relevant to post. Among all the NRM variants

\begin{itemize}[leftmargin=18pt,topsep=1pt]
\setlength \itemsep{-0.1em}
  \item NRM-loc outperforms NRM-glo, suggesting that a dynamically generated context might be more effective than a ``static" fixed-length vector for the entire post, which is consistent with the observation made in~\cite{bahdanau2014neural} for machine translation;
\item  NRM-hyp outperforms NRM-loc and NRM-glo, suggesting that a global representation of post is complementary to dynamically generated local context.
\end{itemize}

The retrieval-based model has the similar mean score as NRM-glo, and its ratio on neutral cases outperforms all the other methods. This is because 1)  the responses retrieved by retrieval-based method are actually wrote by human, so they do not suffer from grammatical and fluency problems, and 2) the combination of various feature functions potentially makes sure the picked responses semantically relevant to test posts. However the picked responses are not customized for new test posts, so the ratio of suitable cases is lower than the three neural generation models.

To test statistical significance, we use the Friedman test~\cite{howell2010fundamental}, which is a non-parametric test on the differences of several related samples, based on ranking. Table~\ref{tab:result_compare} shows the average rankings over all annotations and the corresponding $p$-values for comparisons between different pairs of methods. The comparison between retrieval-based and NRM-glo is not significant and their difference in ranking is tiny. This indicates that the retrieval-based method is comparable to the NRM-glo method. The NRM-hyb outperforms all the other methods, and the difference is statistically significant ($p < 0.05$). The difference between NRM-loc and retrieval-based method is marginal ($p = 0.062$)\comments{\color{blue} [be careful with the terms]}. SMT is significantly worse than retrieval-based and NRM-hyb methods. 

\subsection{Case Study}\label{s:case} 
Figure~\ref{fig:response_examples} shows some example responses generated by our NRMs (only the one with biggest likelihood is given) and the comparable retrieval-based model.  It is intriguing to notice that three NRM variants give suitable but quite distinct responses, with different perspectives and choices of words. This, as we conjecture, is caused by both the architecture variations among models as well as the variations from random effects like the initialization of parameters. Another interesting observation is on the forth example, where the retrieval-based method returns a response with the a mismatched entity name  ``WenShan'', which is actually a quite common problem for retrieval-based model, where the inconsistency details (e.g., dates, named entities), which often render the response unsuitable, cannot be adequately considered in the matching function employed in retrieving the responses. In contrast, we observe that NRMs tend to make general response and barely generate those details.

\begin{figure}[tbh]
\centering
\includegraphics[width=0.5\linewidth]{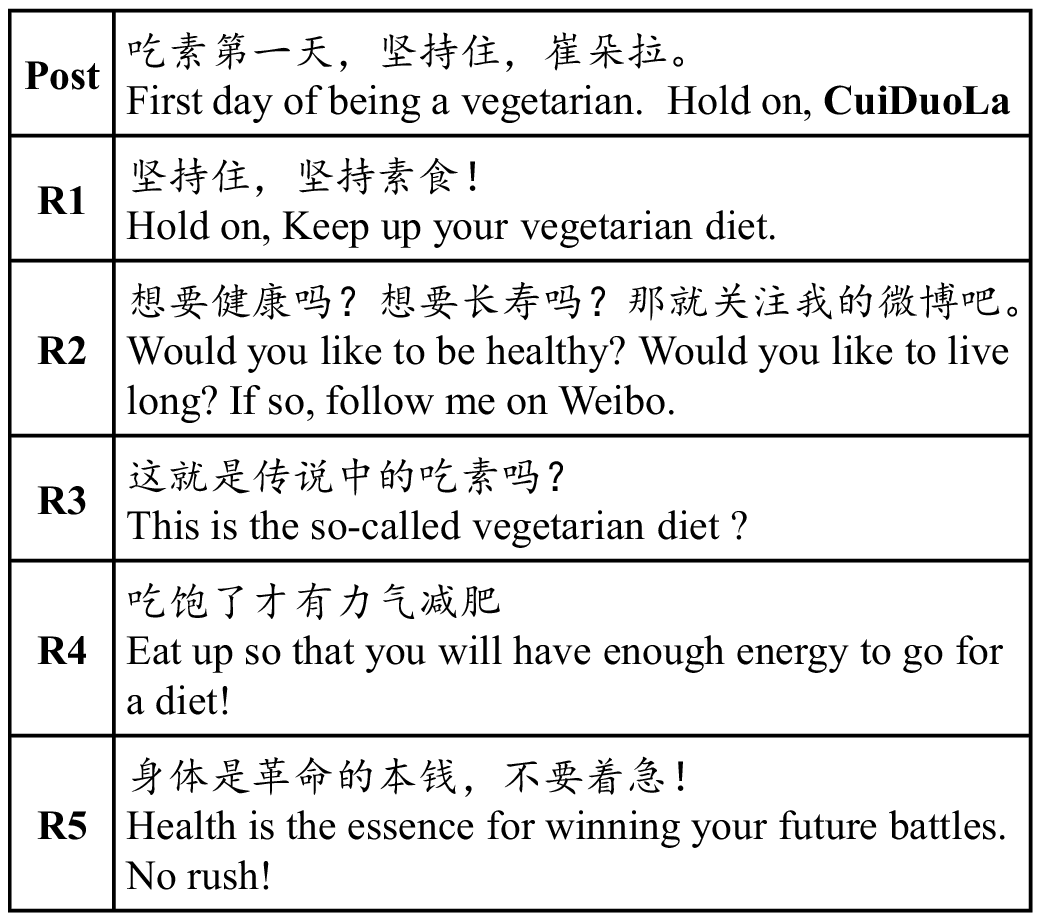}
\caption{Multiple responses generated by the NRM-hyb.} \label{fig:mr}
\end{figure}

We also use the NRM-hyb as an example to investigate the ability of NRM to generate multiple responses. Figure~\ref{fig:mr} lists 5 responses to the same post, which are gotten with beam search with beam size = 500, among which we keep only the best one (biggest likelihood) for each first word.  It can be seen that the responses are fluent, relevant to the post, and still vastly different from each other, validating our initial conjecture that NRM, when fueled with large and rich training corpus,  could work as a generator that can cover a lot of modes in its density estimation. 

\section{Conclusions and Future Work}\label{s:related} 
In this paper, we explored using encoder-decoder-based neural network system, with coined name Neural Responding Machine, to generate responses to a post. Empirical studies confirm that
the newly proposed NRMs, especially the hybrid encoding scheme, can outperform state-of-the-art retrieval-based and SMT-based methods. Our preliminary study also shows that NRM can generate multiple responses with great variety to a given post. In future work, we would consider adding the intention (or sentiment) of users as an external signal of decoder to generate responses with specific goals.

\comments{In this paper, three neural responding machines with different encoding schemas were exploited for generating responses to a post. Empirical studies confirm that
the newly proposed NRM with hybrid scheme has the best performance over the other NRMs and even the state-of-the-art IR-based and SMT-based methods. Our preliminary study also shows that the hybrid NRM can generate multiple responses with various perspectives. In future work, it would be interesting to model the intention (or sentiment) of users as an external signal of decoder to generate responses with specific goals.
How to well combine the IR-based method with generation-based method is also an open and import future direction.}

\bibliographystyle{acl}
\bibliography{acl2015}

\end{document}